\documentclass[letterpaper, 10 pt, conference]{ieeeconf}  % Comment this line out if you need a4paper

\IEEEoverridecommandlockouts                              % This command is only needed if 
\usepackage{cite}
\usepackage{graphicx}
\usepackage{algorithm} 
\usepackage{algpseudocode} 
\usepackage{amsmath}
\usepackage{amssymb}
\usepackage{nicefrac}
\usepackage{caption}
\usepackage{subcaption}
\usepackage{multirow}
\usepackage{tabularx}
\usepackage{siunitx}
\usepackage{booktabs}

\title{\LARGE \bf
Material-agnostic Shaping of Granular Materials with Optimal Transport
}

\author{Nikhilesh Alatur,  Olov Andersson, Roland Siegwart and Lionel Ott% <-this % stops a space
\thanks{*This project has received funding from the
European Union’s Horizon 2020
research and innovation program under
grant agreement No 955356}% <-this % stops a space
\thanks{All authors are members of the Autonomous Systems Lab, ETH Zurich, Switzerland; {\tt\small \{firstname.lastname\}@mavt.ethz.ch}}%
}

\begin{document}
\maketitle
\thispagestyle{empty}
\pagestyle{empty}

%%%%%%%%%%%%%%%%%%%%%%%%%%%%%%%%%%%%%%%%%%%%%%%%%%%%%%%%%%%%%%%%%%%%%%%%%%%%%%%%
\begin{abstract}
    From construction materials, such as sand or asphalt, to kitchen ingredients, like rice, sugar, or salt; the world is full of granular materials. Despite impressive progress in robotic manipulation, manipulating and interacting with granular material remains a challenge due to difficulties in perceiving, representing, modelling, and planning for these variable materials that have complex internal dynamics. While some prior work has looked into estimating or learning accurate dynamics models for granular materials, the literature is still missing a more abstract planning method that can be used for planning manipulation actions for granular materials with unknown material properties. In this work, we leverage tools from optimal transport and connect them to robot motion planning. We propose a heuristics-based sweep planner that does not require knowledge of the material's properties and directly uses a height map representation to generate promising sweeps. These sweeps transform granular material from arbitrary start shapes into arbitrary target shapes. We apply the sweep planner in a fast and reactive feedback loop and avoid the need for model-based planning over multiple time steps. We validate our approach with a large set of simulation and hardware experiments where we show that our method is capable of efficiently solving several complex tasks, including gathering, separating, and shaping of several types of granular materials into different target shapes.
\end{abstract}

%%%%%%%%%%%%%%%%%%%%%%%%%%%%%%%%%%%%%%%%%%%%%%%%%%%%%%%%%%%%%%%%%%%%%%%%%%%%%%%%
\section{INTRODUCTION}
Many real-world tasks require the manipulation of granular materials, such as gravel or grit in construction or rice in cooking.
While robotics has come a long way in manipulating rigid objects, highly deformable materials, including granular materials, still pose a large challenge due to the difficulty of accurately modelling how the material moves in response to a given manipulation action \cite{billard2019trends}.
In their recent survey on manipulating multiple objects, ranging from a few large objects to piles of small objects/particles, Pan et al. \cite{pan2022algorithms} summarise the major challenges that apply to manipulating granular materials as: Perceiving the material of interest, potentially under partial occlusion, efficient state representation of such a high dimensional material, accurate and efficient models for these complex materials that are simple to estimate and plan with, and the need for many different manipulation actions to solve a higher level task, such as scooping, dumping, pushing and so on.

While manipulation of granular materials is still at its infancy, several works have attemped to learn models of materials for planning purposes, or directly learn a particular manipulation action, either with reinforcement learning or learning from demonstration.
While having good models of materials is key for planning precise manipulation actions on a control level, we argue that the state of art in manipulating granular materials could benefit from a higher-level planning layer that can, independently of the actual material being manipulated, plan manipulation actions to achieve a higher level goal in terms of shaping the distribution of the material.

\begin{figure}[bt]
\centering
    \includegraphics[width=0.5\textwidth, trim= 2cm 1cm 2cm 1cm,clip]{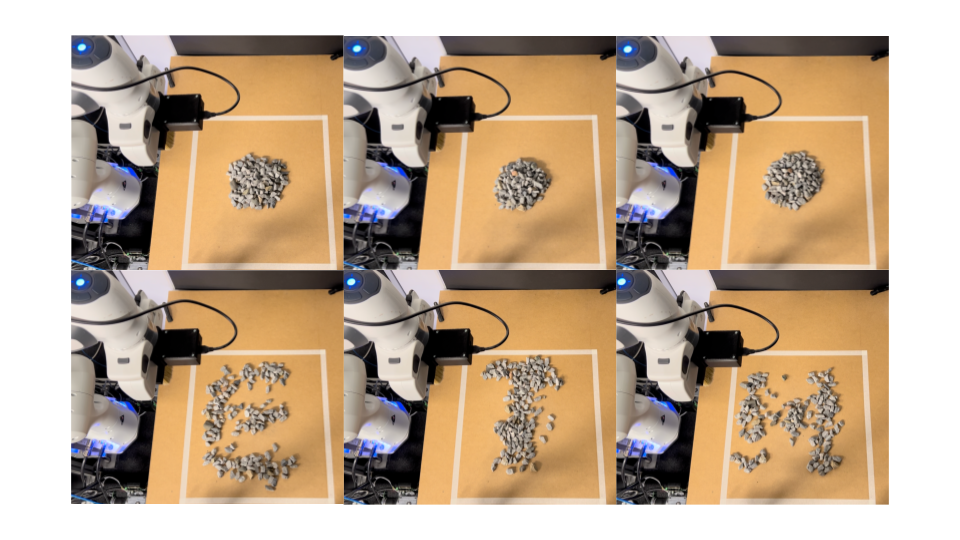}
    \caption{ETH written with granular grit stones using our proposed approach.}
    \label{fig:hw_exp_eth_grit}
\end{figure}

\subsection{Contributions}
Our contributions can be summarised as follows:
In this work, we show that the mathematical theory of Optimal Transport (OT) is a promising framework for planning manipulation actions that require transportation and volumetric shaping of granular materials on a high level, without requiring exact model knowledge of the material.
While OT ignores robot motion constraints, we propose an efficient way of incorporating it into a heuristics-based sweep planner, allowing volumetric shaping for arbitrary input and output distributions. We show that this higher-level shape planner works for a wide variety of tasks and granular materials without requiring an accurate model of material dynamics. Since we can forego computationally expensive models, the proposed approach can run in a reactive feedback loop on a real robot arm using only a height map as input.
We support our claims by an extensive simulation study and real-world experiments for different materials and tasks that require reasoning about how to efficiently re-distribute granular material.

\section{RELATED WORK}
We here categorize related work based on the type and fidelity of the  model used. While an exhaustive survey of modelling approaches for granular dynamics is beyond the scope of this work,  we summarize the findings of \cite{kim2019developing}, \cite{yin2021modeling}, \cite{fang2021chrono}, and \cite{pan2022algorithms}.

\paragraph{First-principle Methods}
Understanding the underlying physics of granular materials has been of great interest to the research community.
The \emph{discrete element method} (DEM) is known to be the most accurate method for modelling granular materials \cite{kim2019developing}, \cite{matl2020inferring},  \cite{hu2021using}, where each grain is modelled as an individual particle that interacts through contact with its neighbouring particles. Unfortunately, due to its particle-level discretization, DEM does not scale well for large volumes and small grain sizes and cannot be run in real-time as suggested by the results in \cite{fang2021chrono} and \cite{hu2021using}.
Furthermore, several hyperparameters such as grain size and shape have to be defined \cite{matl2020inferring}, \cite{hu2021using}, which can be difficult to impossible for real, heterogeneous granular materials encountered in the field.
Matl et al., \cite{matl2020inferring} leveraged the DEM method to perform real-to-sim transfer for pouring granular materials into target shapes with a robotic manipulator. However, as they optimized their model parameters by matching the macroscopic pouring behaviour, it remains unclear how well the estimated model transfers to other manipulation tasks and interaction types.

\paragraph{Simplified Models}
Because of the computational complexity of accurate physics models, many simplified methods have been developed. Position-based dynamics by Macklin and Mueller \cite{macklin2014unified} is one popular method that trades off physical accuracy for visual fidelity and real-time performance \cite{yin2021modeling}. A PBD simulator was used by Zhang et al.~\cite{zhang2020learning} to train an RL agent for manipulating gathering, spreading, and flipping granular and viscous materials. However, no real-world experiments were shown. 
More recently, Tuomainen et al., \cite{tuomainen2022manipulation} train a graph neural network from a simulator that uses the material-point-method for learning the dynamics of pouring actions and use this to plan pouring trajectories that minimizes the Wasserstein distance between the poured and target shape. 
% Howeverlearned dynamics, planning a trajectory takes takes hours. 
Others used task- and action-specific representations, such as Martinez et al. \cite{martinez2015planning} who cluster dirt particles into elliptical regions, learn how these elliptic regions change during pushing actions and then plan trajectories for collecting dirt on a table. Similarly,  Leidner et al.~\cite{leidner2016robotic} employ a hand-crafted heuristic model for a cleaning task, where affected dirt particles are simply placed at the end of each sweep when planning the cleaning trajectories.

\paragraph{Differentiable Physics}
Simulators that represent physical models in a differentiable architecture hold promise for fast parameter estimation and optimization-based planning and control \cite{yin2021modeling}.
Li et al. \cite{li2018learning} proposed the DPI-Net architecture to learn the behaviour of soft objects and liquids from data, in their case generated by the PBD-based FleX simulator from Nvidia.
While this line of work holds a lot of promise, it is still in its infancy \cite{yin2021modeling}, and ready-to-use models that can be quickly adapted to different environments, tasks, and manipulation actions still seem to be missing.

\paragraph{Visual Predictive Models}
Visual predictive models do not attempt to model the granular material in a first-principle fashion, instead, they directly learn how the sensory feedback will change for a given manipulation action.
Schenck et al.~\cite{schenck2017learning} propose a visual-predictive model for scooping pinto beans from a source container and dumping them into an empty target container for different target shapes, where the model is learned from real-world training examples. Elliott et al.~\cite{elliott2018robotic} learn a visual-predictive model, based on pixel-level classifiers, for predicting the outcome of a push-motion to dirt lying on a table and then use A* to plan sweeping trajectories for cleaning the table.
Suh et al. \cite{suh2020surprising} investigate the impact of the network architecture on the quality of the learned visual predictive models. They find that simple linear predictive models outperform deeper neural networks, and demonstrate their model in an impressive experiment where they can gather small carrot pieces into letter-shaped target regions.
Tsuruta et al.~\cite{tsuruta2022deep} train a GAN to predict future sand state (elevation map) when given an input elevation map and a tool path (perpendicular spatula). 
By directly working in the space of the sensory input, visual predictive models avoid the difficult problem of estimating the state of the granular material and also offer an intuitive interface to define the environment and target shape.
However, as they are learned for one specific material-environment-action combination it remains unclear how well these models generalize to different materials and actions or unseen environments.
Also, in the case of image-based models, it is often difficult to represent more complex material distributions that go beyond binary state representations (no-material/material).

\paragraph{Model-free Methods}
Besides model-based methods, there have been several works that have shown impressive real-world results without requiring any type of model.
One line of work uses learning from demonstration, such as in \cite{kim2018icub}, \cite{cauli2018autonomous}, \cite{zeng2020transporter}, or \cite{cherubini2020model} to solve gathering, cleaning, and shaping tasks that involve granular materials.
While these methods can be well suited for repetitive tasks in known environments, they have not been shown to generalize to new environments, materials, or tasks.
Finally, Jud et al. \cite{jud2017planning} use a heuristics-based planner for digging and dumping with an excavator, it is able to execute digging cycles until the currently seen elevation map approximates the target elevation map of a digging site.

\section{METHOD}
\subsection{Overview}
Here we present our framework for generating robot motion plans to transform piles of granular material from arbitrary source shapes into arbitrary target shapes via sweeping motions, without assuming access to material properties or models.
To this end, we adopt an \emph{Eulerian} viewpoint, where the granular material is represented by its volume distribution, in our case by a discrete height map (see \cite{pan2022algorithms} for the difference between Lagrangian and Eulerian state representations).
We leverage discrete optimal transport, a mathematical framework that provides tools for computing distances and transformation maps between arbitrary probability distributions, providing a way to morph a source distribution into a target distribution with the minimum amount of work. 
We then propose a heuristics-based sweep planner for bridging the gap between optimal transport and the motion constraints of robot sweep.

\subsection{Computational Discrete Optimal Transport}
We here give a brief introduction to computational optimal transport and motivate its applicability to plan manipulation actions for transporting granular materials.
Quoting from Solomon's survey on optimal transport on discrete domains \cite{solomon2018optimal}, ``the optimal transport (or Monge–Kantorovich) problem involves the matching of probability distributions defined over a geometric domain like a surface or manifold''.
In the following, we introduce a notation that is specific to our use case of transporting granular materials, adapted from the more general formulation in \cite{solomon2018optimal}.

\paragraph{Discretization into height maps}
We represent the volumetric distribution of the granular material in the form of a discrete height map $H =(h, X)$, which has $N$ entries and where $X_i$ denotes the 2D location of the centre of cell $i$ and $h_i$ denotes the (average) height of the granular material in cell $i$.  

\paragraph{Normalization into probability distribution}
As optimal transport theory is formulated for probability distributions, we introduce a normalized height map $\tilde{h} = \frac{h}{\sum_{i=1}^{(N)}{h_i}}$, and similarly $\tilde{H}$.
\paragraph{Source and target distributions}
The goal of optimal transport is to compute the optimal transport solution to move the "probability mass" from a source into a target distribution.
Hence, we will distinguish two height maps/distributions:
The current state of the granular material, before applying any action, is denoted by $H_\mathcal{S}$ and the desired target height map as $H_\mathcal{T}$.
\paragraph{Ground Cost}
OT finds the optimal way to transport probability mass from $\tilde{H}_\mathcal{S}$ into $\tilde{H}_\mathcal{T}$, where \emph{optimality} is defined with respect to the ground cost, which is a distance metric on the geometric domain on which the distribution is defined.
The ground cost is captured by the matrix $C$ whose entries $c_{ij}$ represent the cost required to transport one unit of probabilistic mass from $\tilde{H}_{\mathcal{S},i}$ to $\tilde{H}_{\mathcal{T},j}$.

\paragraph{Transport Map}
The transport map $T \in \mathbb{R}^{N_\mathcal{S} \times N_\mathcal{T}}$ denotes a transport plan for how the probability mass in $\tilde{H}_\mathcal{S}$ should be re-distributed to $\tilde{H}_\mathcal{T}$.
The entry $T_{ij}$ tells us how much (probability) mass needs to be moved from $\tilde{H}_{\mathcal{S},i}$ to $\tilde{H}_{\mathcal{T},j}$.
Additionally, we require that mass cannot be destroyed or created, only moved:
\begin{equation}
    \begin{cases}
        \sum_j{T_{ij}} = \tilde{h}_{\mathcal{S},i} \\
        \sum_i{T_{ij}} = \tilde{h}_{\mathcal{T},j}
    \end{cases}
\end{equation}

\paragraph{Solution as Linear Program (LP)}
The problem of transporting the probability mass $\tilde{h}_{src}$ to $\tilde{h}_{tgt}$ that is optimal w.r.t. the ground cost matrix $C$ can then be efficiently solved as a linear program, as shown in \cite{solomon2018optimal}:
\begin{equation}
  OT(\tilde{h}_\mathcal{S}, \tilde{h}_\mathcal{T}; C) =
    \begin{cases}
      \text{min}_T \quad \sum_{ij}{(T_{ij} \, c_{ij})} \\
      \quad \text{s.t.} \quad T \geq 0 \\
      \quad \quad \quad \sum_j{T_{ij}} = \tilde{h}_{\mathcal{S},i} \quad \forall i\\
      \quad \quad \quad \sum_i{T_{ij}} = \tilde{h}_{\mathcal{T},j} \quad \forall j
    \end{cases}       
\end{equation}

By solving this linear program, we get the optimal \emph{transport map} $T^*$ as well as the optimal transportation cost $OT(\tilde{h}_\mathcal{S}, \tilde{h}_\mathcal{T}; C) = \sum_{ij}{(T^*_{ij} \, c_{ij})}$.

\paragraph{Choice of Ground Cost}
A common choice for the ground cost is as follows:
\begin{equation}
    c_{ij,p} = \lVert X_{\mathcal{S},i} - X_{\mathcal{T},j} \rVert^{1/p}
\end{equation}
Using this ground cost, induces the so-called $p$-Wasserstein distance \cite{solomon2018optimal}:
\begin{equation}
    \mathcal{W}_p(\tilde{h}_{\mathcal{S}}, \tilde{h}_{\mathcal{T}}) = (OT(\tilde{h}_\mathcal{S}, \tilde{h}_\mathcal{T}; C_p))^{\nicefrac{1}{p}}
\end{equation}

In this work, we use the 1-Wasserstein distance.
Interestingly, the Wasserstein distance is also known as \emph{Earth mover's} distance \cite{rubner2000earth} in the computer vision literature, as the problem of moving earth is often used as an example to introduce the concept of optimal transport and it also hints at its roots in the very first formalization of optimal transport by Monge \cite{monge1781memoire}.
Hence, from now on we will use the term Earth Mover's Distance (EMD) as it fits the theme of this work.

\subsection{Connecting OT with Robot Motion Planning}
While OT at a glance seems very suitable for planning robot motions that require manipulating granular materials, it has seen very little practical application in this field.
Tuomainen et al., \cite{tuomainen2022manipulation} use the Earth Mover's Distance in their robotic pouring problem as an error metric for the poured granular material. However, instead of exploiting the easily computable optimal transport map, they directly optimize on a learned material model, which takes hours for each motion. 
Schenck et al.~\cite{schenck2017learning} mention that the earth mover's distance as a possible error metric in their scooping and dumping problem for granular material, but instead fall back to the L1 norm due to its computational simplicity.
One major problem is that Optimal Transport assumes that the (probability) mass in each cell can be split up into infinitesimally small chunks and moved individually to other cells without any extra cost.
While this might approximately hold for highly localized actions like precise scooping and dumping, this is not the case for manipulation actions, such as sweeping, that affect a large region of the granular material distribution.
In the following, we leverage not only the error metric provided by OT, but also the transport map $T*$ for planning efficient robot motion that requires transporting granular materials via sweeping motions of a robot arm.

\subsection{Assumptions}
In the following, we will assume that the granular material of interest is incompressible, has a grain size that is much smaller than the end-effector, and have negligible cohesive and adhesive effects which enables the use of position-controlled sweeping trajectories.
We will further assume that the material is (and will remain) on a planar, gravity-aligned, convex, and obstacle-free workspace that is completely reachable by the robotic end-effector.
Lastly, we assume that the volume distribution of the granular material is given in the form an occlusion-free, discrete height map.

\subsection{Next-Best-Sweep Planner}
\begin{figure}[bt]
\centering
    \includegraphics[width=0.5\textwidth, trim= 8cm 1cm 8cm 1cm,clip]{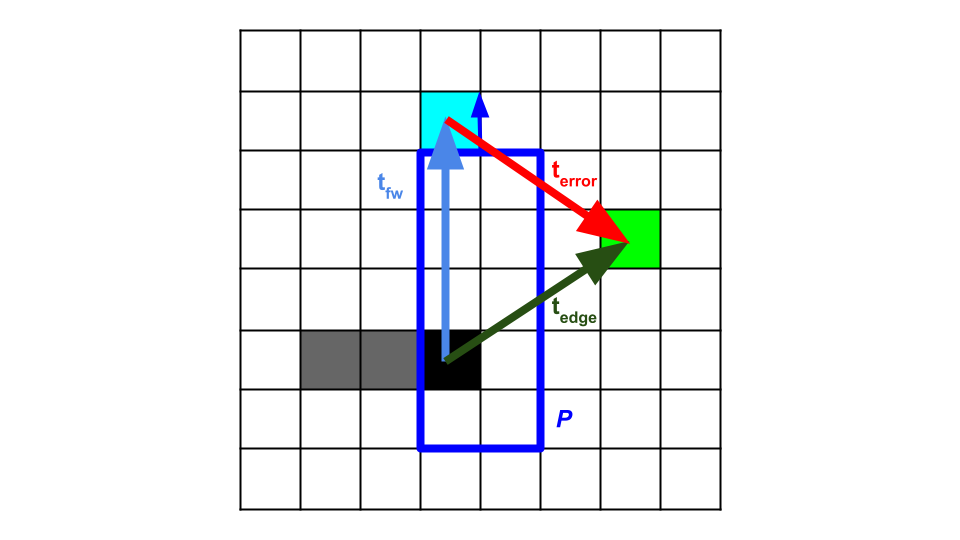}
    \caption{Simple forward push model used for computing the sweep score.}
    \label{fig:sweep_score}
\end{figure}

To avoid computationally prohibitive sweep planning based on accurate dynamics models of the materials, we instead propose to use a next-best sweep planner that computes promising sweep candidates based on the optimal transport map $T*$ between the current heightmap $\tilde{H}_\mathcal{S}$ and the target height map $\tilde{H}_\mathcal{T}$.
We will first introduce a heuristic scoring function $g(T, a)$ that takes as input the transport map $T$ and a sweep primitive $a$ to compute a proxy score on how well the sweep agrees with the transport map.
We will use $g(T, a)$ to rank a set of $L$ candidate sweeps $\mathcal{A}_{sweep}=\{a_1,...,a_L\}$ and choose the sweep $a^*$ with the best score such that $g(T, a^*) \geq g(T,a_i), \; \forall a_i \in \mathcal{A}_{sweep}$.
\paragraph{Sweep Action Primitive}
We will define the sweep action primitive $a(X_{start}, X_{end}; w_{spatula})$ as a straight line sweep with a perfectly perpendicular and plane-parallel spatula, parameterized by the sweep start location $X_{start}$, sweep end location $X_{end}$, and the spatula width $w_{spatula}$.
\paragraph{Simple Push Model}
Before defining the proxy score, we will first introduce the simple forward push model, which has been inspired by \cite{leidner2016robotic}, and \cite{suh2020surprising}.
The push model is visually depicted in Figure \ref{fig:sweep_score}, where all cells in the source height map $H_\mathcal{S}$ that fall inside the rectangular sweep patch spanned by $a_{i,start}$, $a_{i,end}$, and $w_{spatula}$, shown as black cell, are predicted to end up at the end of the sweep, shown as the cyan coloured cell.
The sweep patch, called $P$, is shown by the blue, directed rectangle in Figure \ref{fig:sweep_score}.
Further, we introduce the corresponding displacement vector as $\vec{t}_{fw}$ and its normal vector $\vec{n}_{fw} = \vec{t}_{fw}/||\vec{t}_{fw}||$
\paragraph{Proxy Score}
We define a proxy score for a sweep as the sum of the mass $T_{ij}$ that is supposed to be moved according to the computed OT transport map, for all edges starting inside the sweep patch $X_{\mathcal{S},i} \in P$, weighted by a heuristic function $r(i,j)$, 
$$
g(T, a) = \sum_{i=1}^{N_\mathcal{S}} \mathbf{1}_{P}(X_S,_i) \sum_{j=1}^{N_\mathcal{T}} T_{ij}\,r(i,j).
$$
Here $\mathbf{1}_{P}()$ is the indicator function for set P. The heuristic $r(i,j)$ is constructed to reflect how well the simple push-model predictions $\vec{t}_{fw}$ (shown in Fig.~\ref{fig:sweep_score}) for the sweep aligns with the OT vectors $\vec{t}_{edge}$, 
where $$r(i,j) = \alpha_{+} \operatorname{max}(\vec{n}_{fw}\cdot \vec{t}_{edge}, 0) + \alpha_{-} \operatorname{min}(\vec{n}_{fw}\cdot \vec{t}_{error}, 0).$$
Additionally, the heuristic function contains a penalty term for overshoot $(\vec{n}_{fw}\cdot \vec{t}_{error})$ (see Fig.~\ref{fig:sweep_score}) so that material is not pushed further than where it was supposed to go (green cell). Here $\alpha_{+}$ and $\alpha_{-}$ are two scalar, positive hyperparameters tuned by a simple parameter sweep in simulation.

\paragraph{Sampling Sweeps}
We compute the next-best sweep to execute by generating a set of candidate sweeps $\mathcal{A}$ and computing the proxy score $g$ for all of them and choosing the highest-ranking sweep.
For sampling the sweeps to evaluate, we use the transport map $T$ as it gives us a strong prior where to look for sweeps that move a lot of probability mass.
First, we sample an edge pair from the probability distribution induced by $T$ as $(i,j) \sim T$ which we use to define the start of the sweep $X_{start} = X_{\mathcal{S},i}$ and the end of the sweep $X_{end} = X_{\mathcal{T},j}$, while discarding very short sweeps where $||X_{end} - X_{start}|| < \delta_{min}$.
For covering a larger range of potential sweeps, for example for sweeps that start at $X_{start}$ but do not have to go all the way to $X_{end}$, we generate further samples by linearly interpolating the sweep end between $X_{start}$ and $X_{end}$ at a step size of $\delta_{refine}$.
We repeat this sampling $L$ times to generate our set of candidate sweeps $\mathcal{A}$.

\section{EXPERIMENTS AND RESULTS}

In the following we demonstrate the effectiveness of our proposed method. First we show extensive quantitative results in simulation on different tasks, comparing our method against two baselines evaluated using the earth mover's distance. Then we show the ability of our approach to handle varied real-world granular materials by performing multiple shaping tasks with a Panda robot arm.

\subsection{Simulation Setup}
\begin{figure}[]
     \centering
     \begin{subfigure}[b]{0.49\columnwidth}
         \centering
         \includegraphics[width=\columnwidth, trim= 7cm 0cm 7cm 0cm,clip]{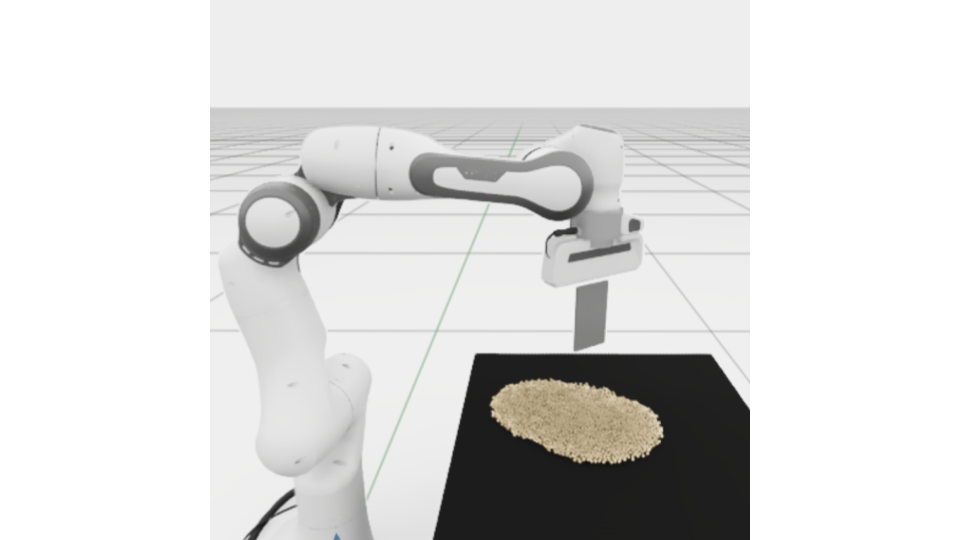}
         \caption{Simulation setup}
         \label{fig:sandpanda_sim}
     \end{subfigure}
     \hfill
     \begin{subfigure}[b]{0.49\columnwidth}
         \centering
         \includegraphics[width=\columnwidth, trim={7cm 0cm 7cm 0cm},clip]{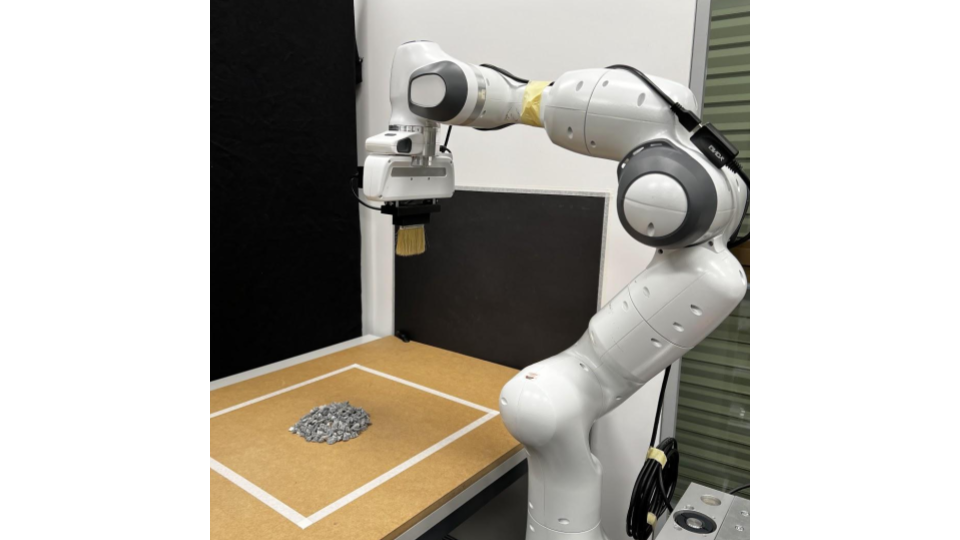}
         \caption{Hardware setup}
         \label{fig:sandpanda_hw}
     \end{subfigure}
      \caption{Experimental setup for a) simulation, and b) hardware experiments.}
\end{figure}

We use Nvidia's Isaac simulator\footnote{https://developer.nvidia.com/isaac-sim} because it supports simulating granular material in real time.
The Isaac simulator leverages the Position-Based Dynamics approach by Macklin et al.~\cite{macklin2014unified} to simulate fluids and granular materials.
The environment consists of a simulated Franka-Emika Panda arm with a \SI{7}{\centi\meter} wide spatula as end-effector, a \SI{0.5}{\meter} $\times$ \SI{0.5}{\meter} large workspace fully reachable by the robot arm.
The depth measurements from the rendered depth camera are converted to a point cloud and fused into a discrete height map at a resolution of \SI{2}{\centi\meter} with the approach from Fankhauser et al. \cite{Fankhauser2014RobotCentricElevationMapping}, \cite{Fankhauser2018ProbabilisticTerrainMapping}.
After computing the straight line sweeps, the motions are executed by feeding way points along the sweep to the RMPFlow controller \cite{cheng2021rmpflow} that ships with Isaac.
See Figure \ref{fig:sandpanda_sim} for a visualization of the simulation setup.
We use the 'Python Optimal Transport' toolbox by Flamary et al., \cite{flamary2021pot} for computing the optimal transport map $T$ and the earth mover's distance.

\subsection{Simulation Tasks}
We will show the usefulness of Optimal Transport as a general tool for planning earth-moving operations and the effectiveness of our proxy-score heuristic planner for pushing granular material into many different shapes.
We will use three different tasks to show the validity of the above claims.

\paragraph{Gathering}
The gathering task is the simplest task, where scattered material should be collected into a single small target region, as shown in \cite{martinez2015planning}, \cite{leidner2016robotic}, \cite{suh2020surprising}, \cite{zeng2020transporter}.
\paragraph{Separation}
Separating granular material into different blobs is more complex than simple gathering, as already shown in \cite{zhang2020learning}. Here the robotic agent needs to reason about how to efficiently re-distribute the granular material from the source distribution into the target distribution.
We differentiate between different separation tasks and denote SEP-N as the task of separation into \emph{N} cluster of equal size.
\paragraph{Letter}
Sweeping granular materials into more complex target shapes is an even more difficult task that requires precise reasoning about the material distribution and planning  the appropriate robot motions.
We choose letters, which was also used in \cite{suh2020surprising} as  examples of complex shapes.

\begin{figure}[bt]
     \centering

    \hfill
    \includegraphics[width=0.185\columnwidth, trim= 17.2cm 1cm 2cm 1cm,clip]{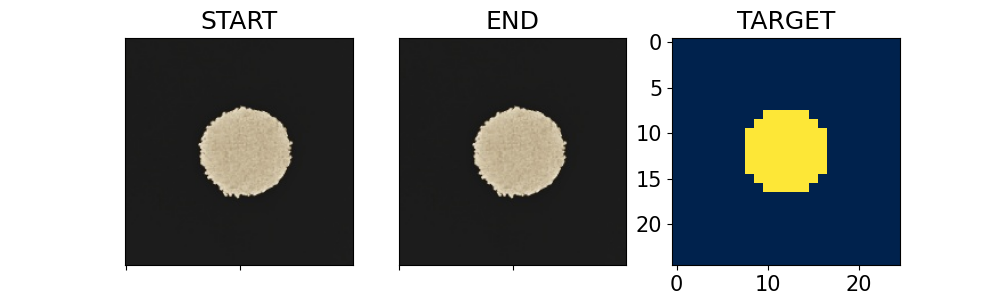}
    \includegraphics[width=0.185\columnwidth, trim= 17.2cm 1cm 2cm 1cm,clip]{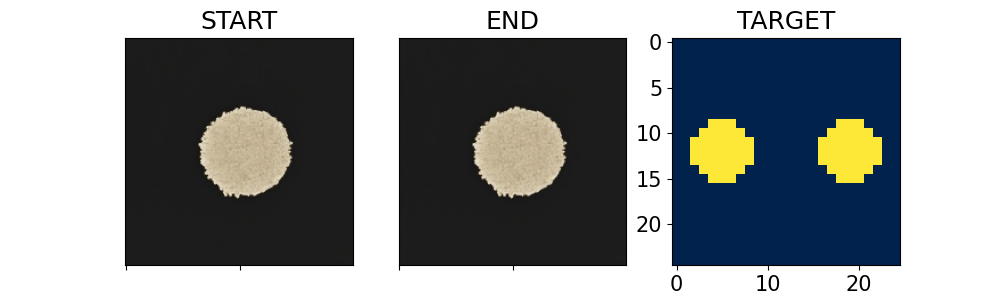}
    \includegraphics[width=0.185\columnwidth, trim= 17.2cm 1cm 2cm 1cm,clip]{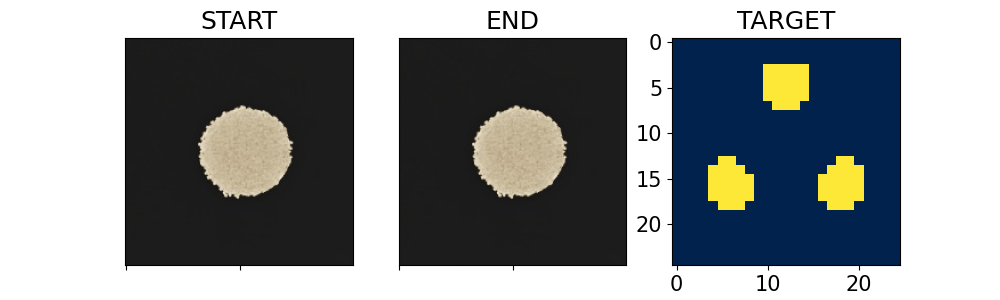}
    \includegraphics[width=0.185\columnwidth, trim= 17.2cm 1cm 2cm 1cm,clip]{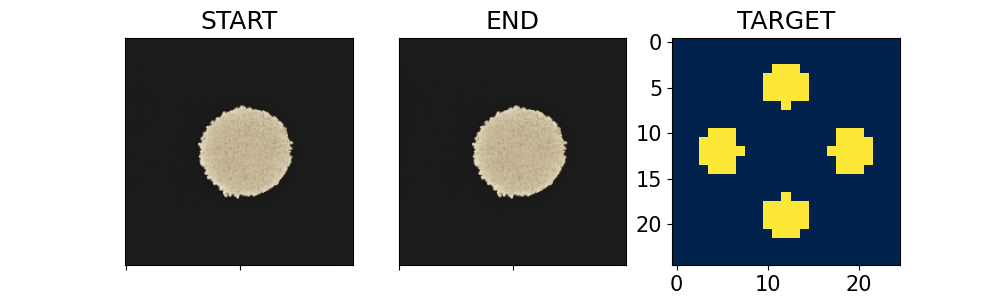}
    \includegraphics[width=0.185\columnwidth, trim= 17.2cm 1cm 2cm 1cm,clip]{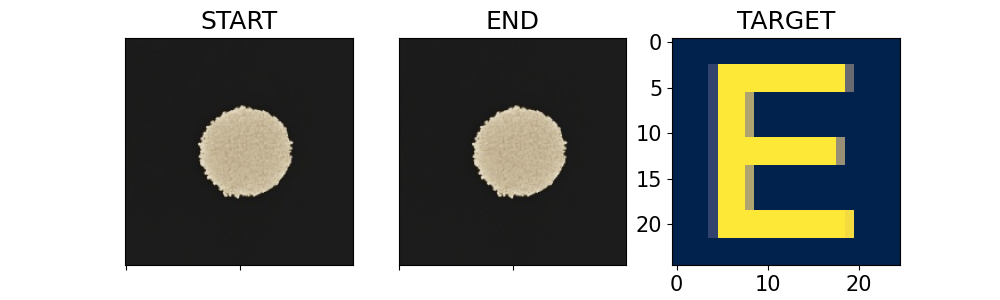}
    \hfill

    \caption{Examples for target shapes: GATHER, SEP-2, SEP-3, SEP-4, LETTER.}
    \label{fig:sim_exp_tgt_shapes}
\end{figure}

\begin{figure}[bt]
     \centering

    \hfill
    \includegraphics[width=0.185\columnwidth, trim= 2cm 1cm 16cm 1cm,clip]{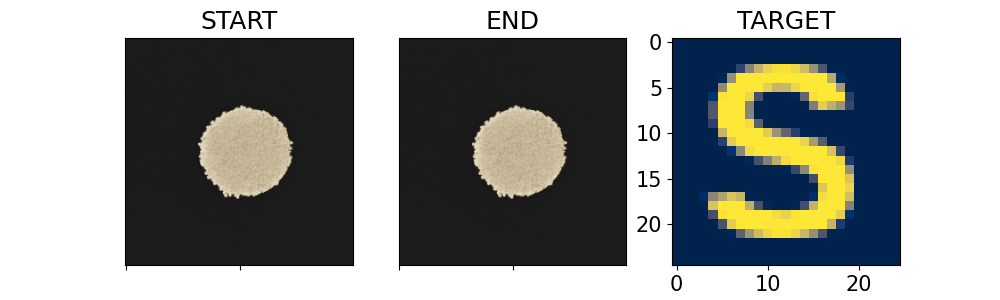}
    \includegraphics[width=0.185\columnwidth, trim= 2cm 1cm 16cm 1cm,clip]{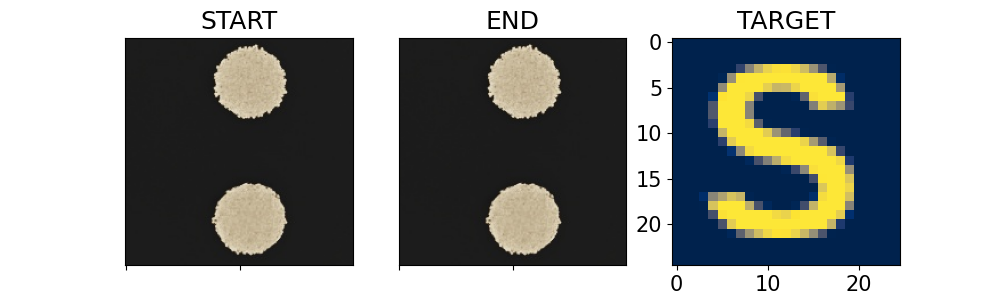}
    \includegraphics[width=0.185\columnwidth, trim= 2cm 1cm 16cm 1cm,clip]{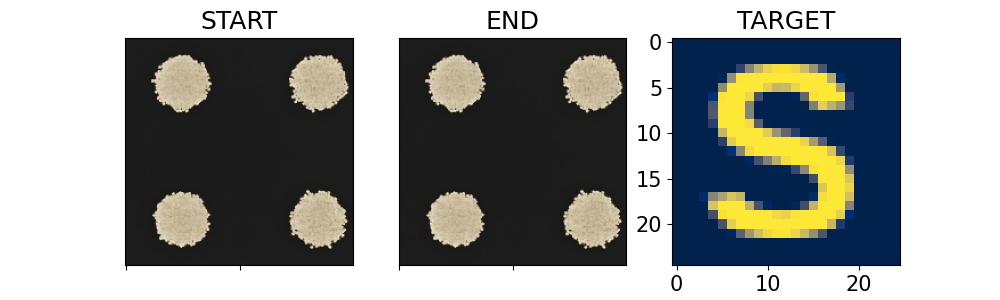}
    \includegraphics[width=0.185\columnwidth, trim= 2cm 1cm 16cm 1cm,clip]{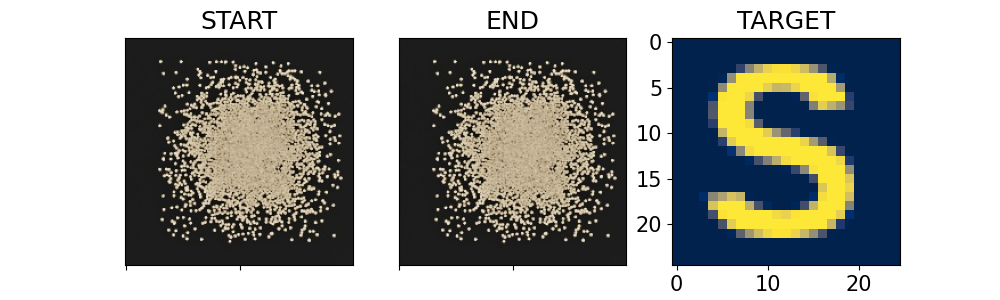}
    \includegraphics[width=0.185\columnwidth, trim= 2cm 1cm 16cm 1cm,clip]{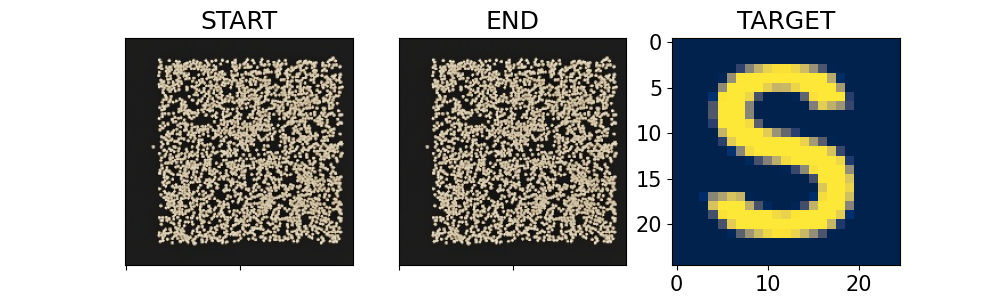}
    \hfill

    \caption{Different initial distributions for the PBD material: 1-blob, 2-blobs, 4-blobs, Gaussian, Uniform.}
    \label{fig:sim_exp_initial_distributions}
\end{figure}

We also use five different initial distributions for the PBD material in the gathering and separation tasks, shown in Figure \ref{fig:sim_exp_initial_distributions}. These vary in terms of scatter across the whole workspace to demonstrate the generalizability of our approach across different initial conditions.

\subsection{Methods}
We will compare our proposed planner against two other baseline methods \emph{MAX-OT} and \emph{DIFF-MAP}, which are briefly summarized in the following:

We will assume that both $\tilde{H}_\mathcal{S}$ and $\tilde{H}_\mathcal{T}$ cover the same area, which is required for the \emph{DIFF-MAP} methods to work.

\paragraph{OURS}
We deploy our proposed OT-based sweep planner with $\alpha_{+}=1.0$, $\alpha_{error}=100.0$, $L_{samples}=10$ and $\delta_{refine}=0.02$.
The high value $\alpha_{-}=100.0$ leads to the planner approaching the target shape more conservatively, resulting in it overshooting less and removing material that was already in place.
The same parameter set was used for all tasks, both in the simulation experiments as well as the physical experiments.

\paragraph{MAX-OT}
The first baseline computes the optimal transport map $T* = OT(\tilde{H}_\mathcal{S}, \tilde{H}_\mathcal{T}, C)$ and then selects the next-best sweep as the strongest non-trivial edge in the transport map $a = (X_{\mathcal{S},i}, X_{\mathcal{T},j})$ where $\{(ij) \mid T_{ij} = max(T) \land X_{\mathcal{S},i} \neq X_{\mathcal{T},j}\}$. This reflects a naive application of OT to the problem that does not respect the motion constraints of a sweep.

\paragraph{DIFF-MAP}
Our second baseline does not use optimal transport at all and is instead inspired by the heuristic planner from \cite{jud2017planning}.
We construct two new height maps $\tilde{H}_{excess}$, where $\tilde{h}_{excess} = max(\tilde{h}_\mathcal{S}-\tilde{h}_\mathcal{T}, 0.0)$ and $\tilde{H}_{lack}$, where $\tilde{h}_{lack} = max(\tilde{h}_\mathcal{T}-\tilde{h}_\mathcal{S}, 0.0)$

We then sample the start of the sweep from the first probability distribution $a_{s,start} \sim \tilde{H}_{excess}$ and the end from the latter $a_{s,end} \sim \tilde{H}_{lack}$

\subsection{Simulation Results}
\begin{figure*}[ht]
\centering
     \begin{subfigure}[b]{0.6\columnwidth}
     \centering
     \includegraphics[width=\columnwidth, trim={0cm 0cm 0cm 0cm},clip]{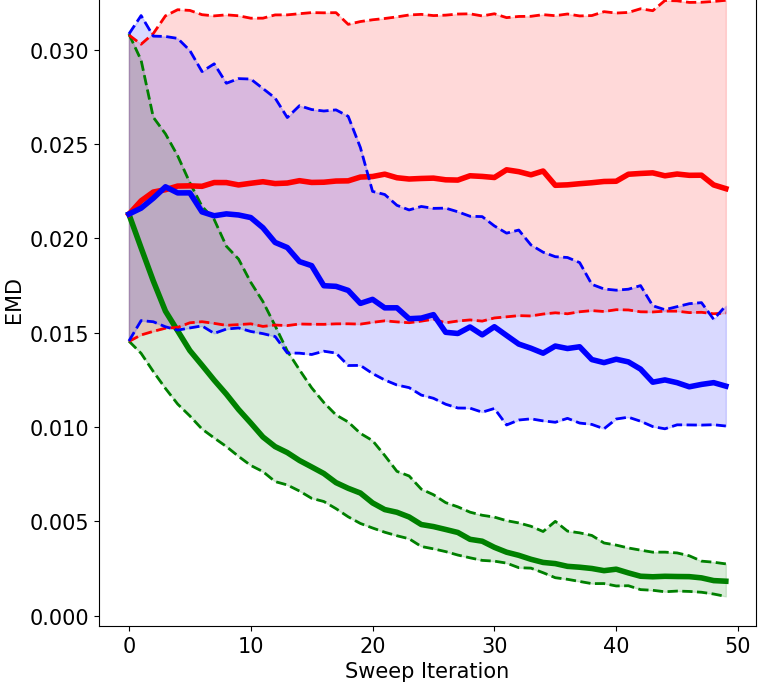}
     \caption{EMD: Gathering in \textbf{Uniform}}
     \label{fig:sim_exp_gather_quantiles_emd}
     \end{subfigure}
     \hfill
     \begin{subfigure}[b]{0.6\columnwidth}
         \centering
         \includegraphics[width=\columnwidth, trim={0cm 0cm 0cm 0cm},clip]{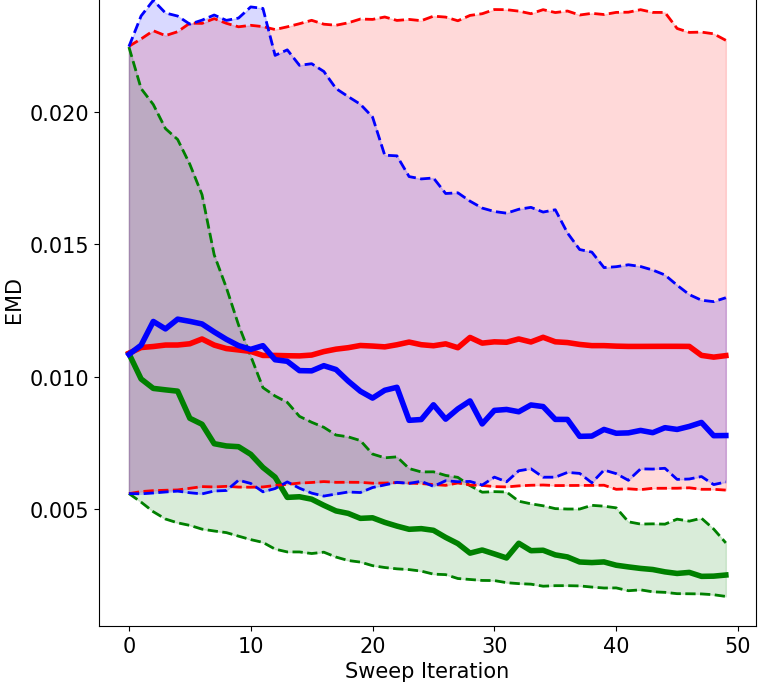}
         \caption{EMD: Separation in \textbf{Uniform}}
         \label{fig:sim_exp_separation_quantiles_emd}
     \end{subfigure}
      \hfill
     \begin{subfigure}[b]{0.6\columnwidth}
         \centering
         \includegraphics[width=\columnwidth, trim={0cm 0cm 0cm 0cm},clip]{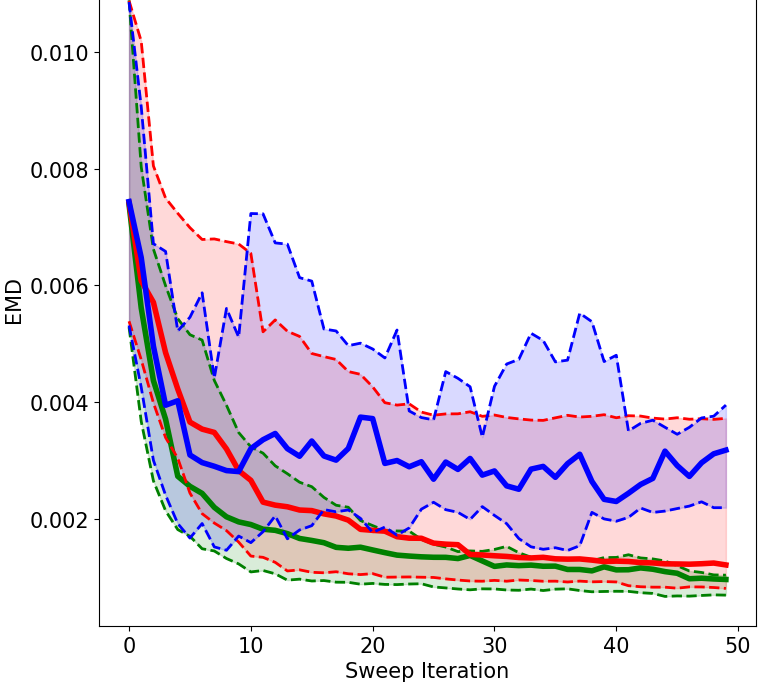}
         \caption{EMD: Letter in \textbf{1-blob}}
         \label{fig:sim_exp_shaping_quantiles_emd}
     \end{subfigure}
    % separate
         \begin{subfigure}[b]{0.6\columnwidth}
     \centering
     \includegraphics[width=\columnwidth, trim={0cm 0cm 0cm 0cm},clip]{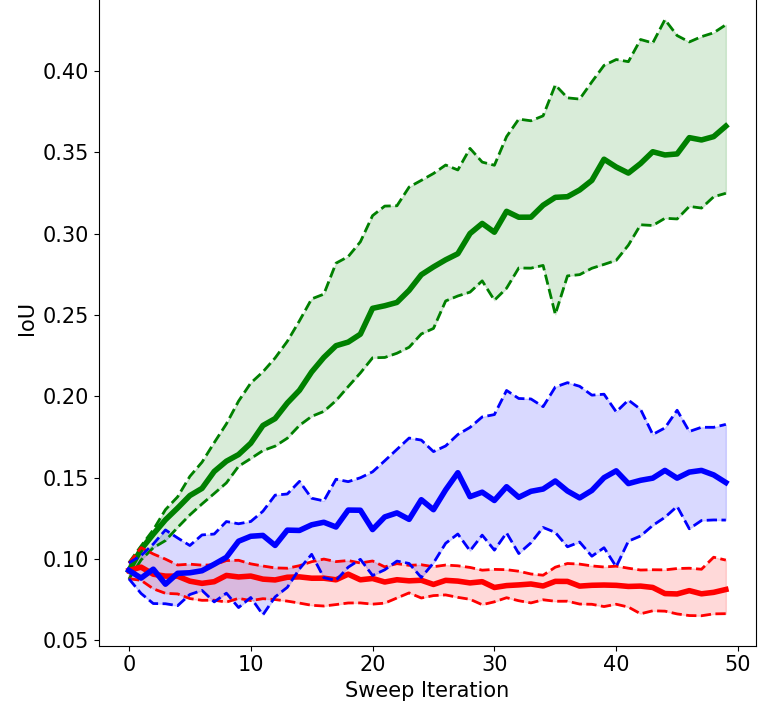}
     \caption{IoU: Gathering in \textbf{Uniform}}
     \label{fig:sim_exp_gather_quantiles_iou}
     \end{subfigure}
     \hfill
     \begin{subfigure}[b]{0.6\columnwidth}
         \centering
         \includegraphics[width=\columnwidth, trim={0cm 0cm 0cm 0cm},clip]{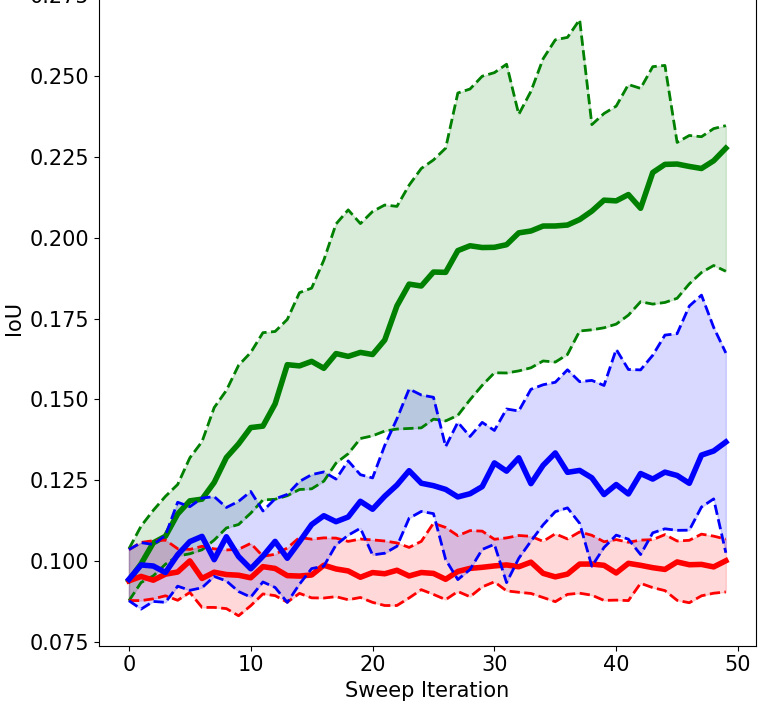}
         \caption{IoU: Separation in \textbf{Uniform}}
         \label{fig:sim_exp_separation_quantiles_iou}
     \end{subfigure}
      \hfill
     \begin{subfigure}[b]{0.6\columnwidth}
         \centering
         \includegraphics[width=\columnwidth, trim={0cm 0cm 0cm 0cm},clip]{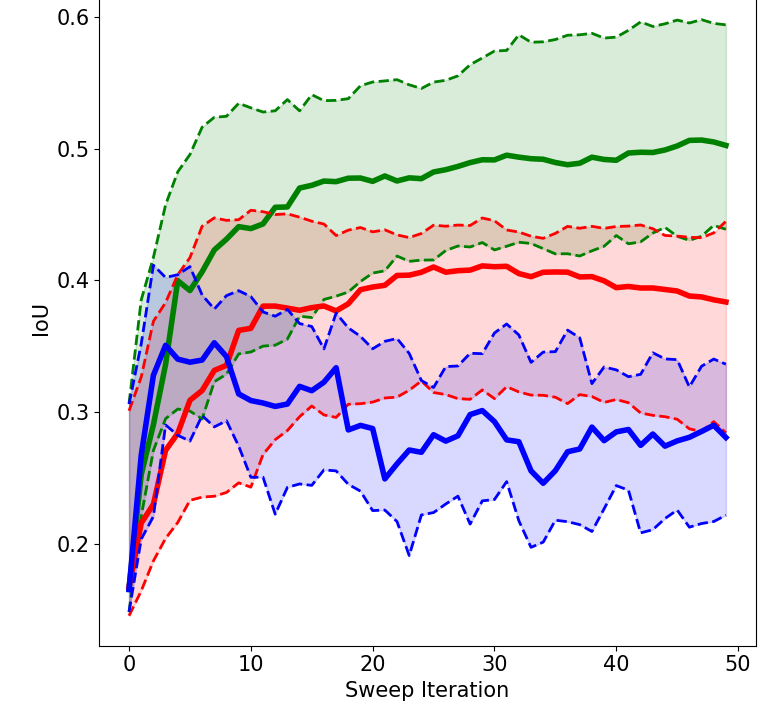}
         \caption{IoU: Letter in \textbf{1-blob}}
         \label{fig:sim_exp_shaping_quantiles_iou}
     \end{subfigure}
    \caption{5\%/50\%/95\% quantiles for selected simulation runs (green: OURS, red: MAX-OT, blue: DIFF-MAP).}
    \label{fig:sim_task_overview}
\end{figure*}

We generated 10 randomized gathering targets by randomly choosing the centre of the circle and 9 randomized separation targets (three each for SEP-2/SEP-3/SEP-4) in a similar fashion.
We let each of the methods perform for 30 iterations on the simpler start environments (\textbf{1-blob}, \textbf{2-blobs}, \textbf{Gaussian}) and for 50 iterations on the more challenging (more scattered) environments (\textbf{4-blobs}, \textbf{Uniform}).
In total we let each method solve 95 different sweeping tasks, each with a different start and end configuration.
In addition, we ran six runs with 50 sweeps in the \textbf{1-blob} environment for shaping the letters "E, T, H, A, S, L". 
The results for the gathering tasks are summarized in Table \ref{tab:gather} and the results for separation are summarized in Table \ref{tab:separation} and they are separated by the chosen initial condition.
The reported values are the earth mover's distance scaled up by $10^{-3}$, given as the median and the $5\%$ and $95\%$ quantiles in the subsequent brackets.
We see that our method consistently outperforms the two other baseline methods, with only a few cases where the $95\%$ quantiles overlap with the $5\%$ quantile of one of the baseline methods. The proposed method also  consistently has the lowest variance by a large margin.
This becomes obvious when looking at the evolution of the metrics over time in Figure \ref{fig:sim_task_overview}, where the median, $5\%$ quantile, and $95\%$ quantile curves are plotted for OUR method and the two baselines for gathering in \textbf{Uniform}, separation in \textbf{Uniform}, and letter in \textbf{1-blob}.
The plots show that our method manages to quickly decrease the EMD values to a low value with a small variance.
We have also plotted the Intersection-over-Union to provide a second metric, that is not based on optimal transport.

\begin{table}[]
    \centering
    \begin{tabular}{lrrrr}
        \toprule
        & INITIAL & OURS & MAX-OT & DIFF-MAP \\
        \midrule
        \textbf{1-blob} & 2.9 & \textbf{1.2} & 2.7 & 3.7 \\
        & [10.6, 23.1] & [1.5, 2.9] & [5.2, 9.5] & [5.4, 8.1] \\

        \textbf{2-blobs} & 15.7 & \textbf{1.8} & 10.0 & 6.9 \\ 
        & [22.8, 32.7] & [2.5, 3.8] & [11.9, 17.8] & [8.9, 13.4] \\ 
        
        \textbf{4-blobs} & 28.6 & \textbf{2.1} & 16.8 & 12.1 \\ 
        & [35.5, 45.3] & [2.6, 3.6] & [22.0, 29.5] & [14.6, 21.1] \\
    \textbf{Gaussian} & 9.7 & \textbf{1.6} & 10.2 & 7.8 \\ 
        & [16.7, 26.7] & [2.2, 3.0] & [15.9, 24.2] & [10.6, 16.5] \\

        \textbf{Uniform} & 14.6 & \textbf{1.0} & 16.0 & 10.0 \\ 
        & [21.3, 30.8] & [1.8, 2.7] & [22.6, 32.6] & [12.2, 16.4] \\ 
        \bottomrule
    \end{tabular}
    \caption{Simulation results for methods (columns) on  gathering tasks, using five different source shapes (rows) and ten random target shapes.}
    \label{tab:gather}
\end{table}

\begin{table}[bt]
    \centering
    \begin{tabular}{lrrrr}
        \toprule
        & INITIAL & OURS & MAX-OT & DIFF-MAP \\
        \midrule
        \textbf{1-blob} & 8.1 & \textbf{1.4} & 3.0 & 3.4 \\
        & [11.2, 22.0] & [2.3, 3.7] & [3.8, 9.0] & [4.0, 5.9] \\
        \textbf{2-blobs} & 5.8 & \textbf{1.9} & 3.5 & 4.5 \\ 
        & [17.4, 33.8] & [3.5, 4.4] & [7.0, 17.3] & [6.4, 13.5] \\ 
        \textbf{4-blobs} & 10.2 & \textbf{1.8} & 6.4 & 6.9 \\ 
        & [18.2, 31.2] & [2.0, 6.5] & [11.1, 20.6] & [10.2, 18.3] \\
        \textbf{Gaussian} & 5.1 & \textbf{2.2} & 4.6 & 4.3 \\ 
        & [10.1, 21.5] & [2.9, 6.0] & [8.6, 18.9] & [6.0, 12.4] \\ 
        \textbf{Uniform} & 5.6 & \textbf{1.7} & 5.7 & 6.0 \\ 
        & [10.8, 22.5] & [2.5, 3.7] & [10.8, 22.7] & [7.8, 13.0] \\ 
    \bottomrule
    \end{tabular}
    \caption{Simulation results for methods (columns) on separation tasks, using five different source shapes (rows) and nine random target shapes.}
    \label{tab:separation}
\end{table}

\subsection{Hardware Setup}
Our setup for the hardware experiment is shown in Figure \ref{fig:sandpanda_hw}.
The setup consists of a Franka-Emika Panda robot arm to which we attach a brush as an end-effector, similar to \cite{elliott2018robotic}, for mechanical compliance when interacting with the rigid ground or rigid particles.
We use the Pico Monstar Time-of-Flight camera from pmdtechnologies to get high-accuracy point cloud scans, which are integrated by the same elevation mapping framework by Fankhauser et al. \cite{Fankhauser2014RobotCentricElevationMapping}, \cite{Fankhauser2018ProbabilisticTerrainMapping} as for the simulation experiments.
We use a fixed scanning position that yields a complete scan of the workspace. The PILZ Industrial Motion Planner, which ships with the popular \emph{MoveIt}\footnote{https://moveit.ros.org/} planning framework, is used to move the end effector on smooth linear trajectories.

\begin{figure}[bt]
     \centering
     \begin{subfigure}[b]{0.32\columnwidth}
         \centering
         \includegraphics[width=\columnwidth, trim= 5cm 5cm 5cm 2cm,clip]{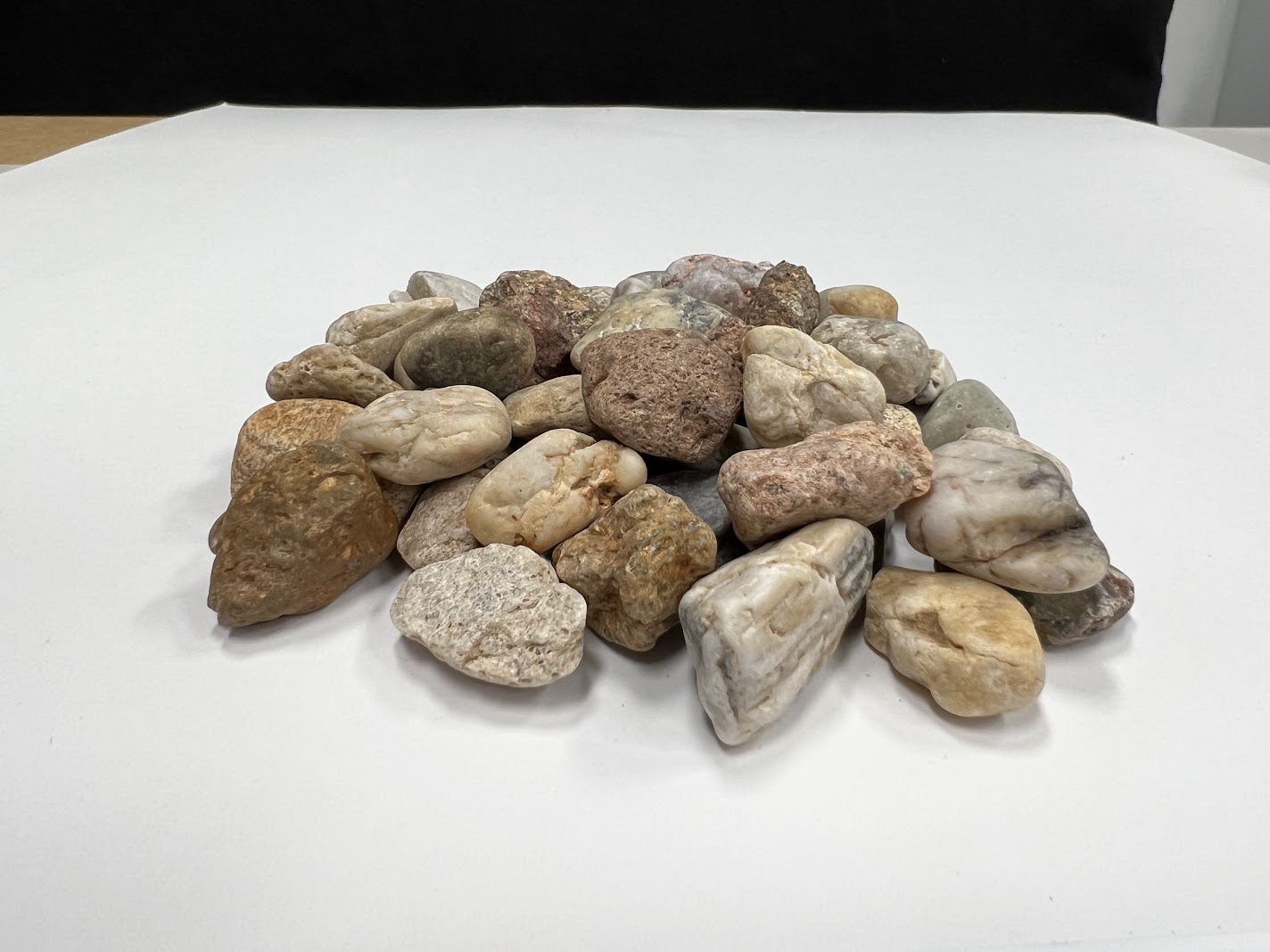}
         \caption{Grind}
         \label{fig:hw_exp_grind}
     \end{subfigure}
     \hfill
     \begin{subfigure}[b]{0.32\columnwidth}
         \centering
         \includegraphics[width=\columnwidth, trim={5cm 5cm 5cm 2cm},clip]{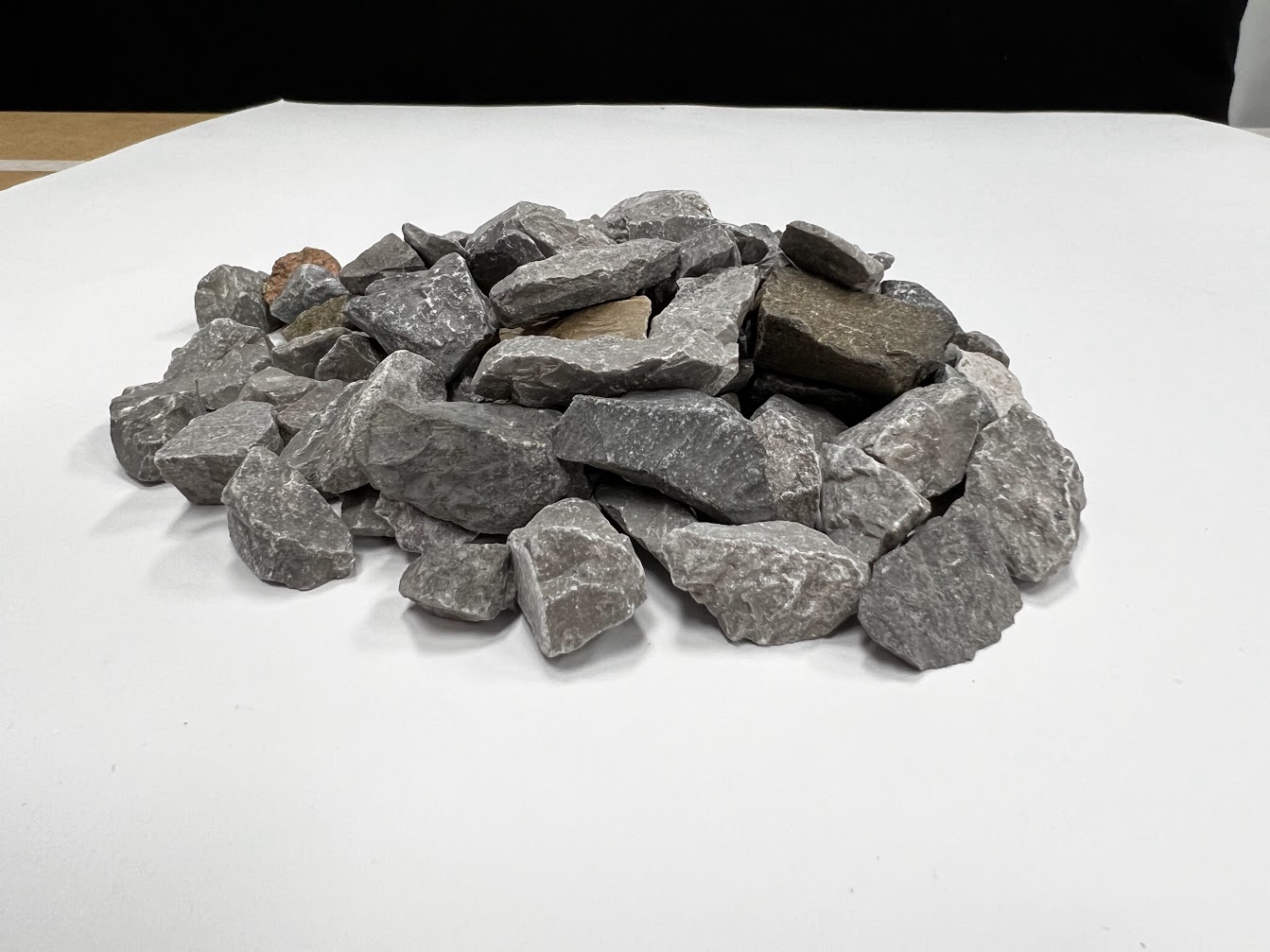}
         \caption{Grit}
         \label{fig:hw_exp_grit}
     \end{subfigure}
     \hfill
     \begin{subfigure}[b]{0.32\columnwidth}
         \centering
         \includegraphics[width=\columnwidth, trim={5cm 5cm 5cm 2cm},clip]{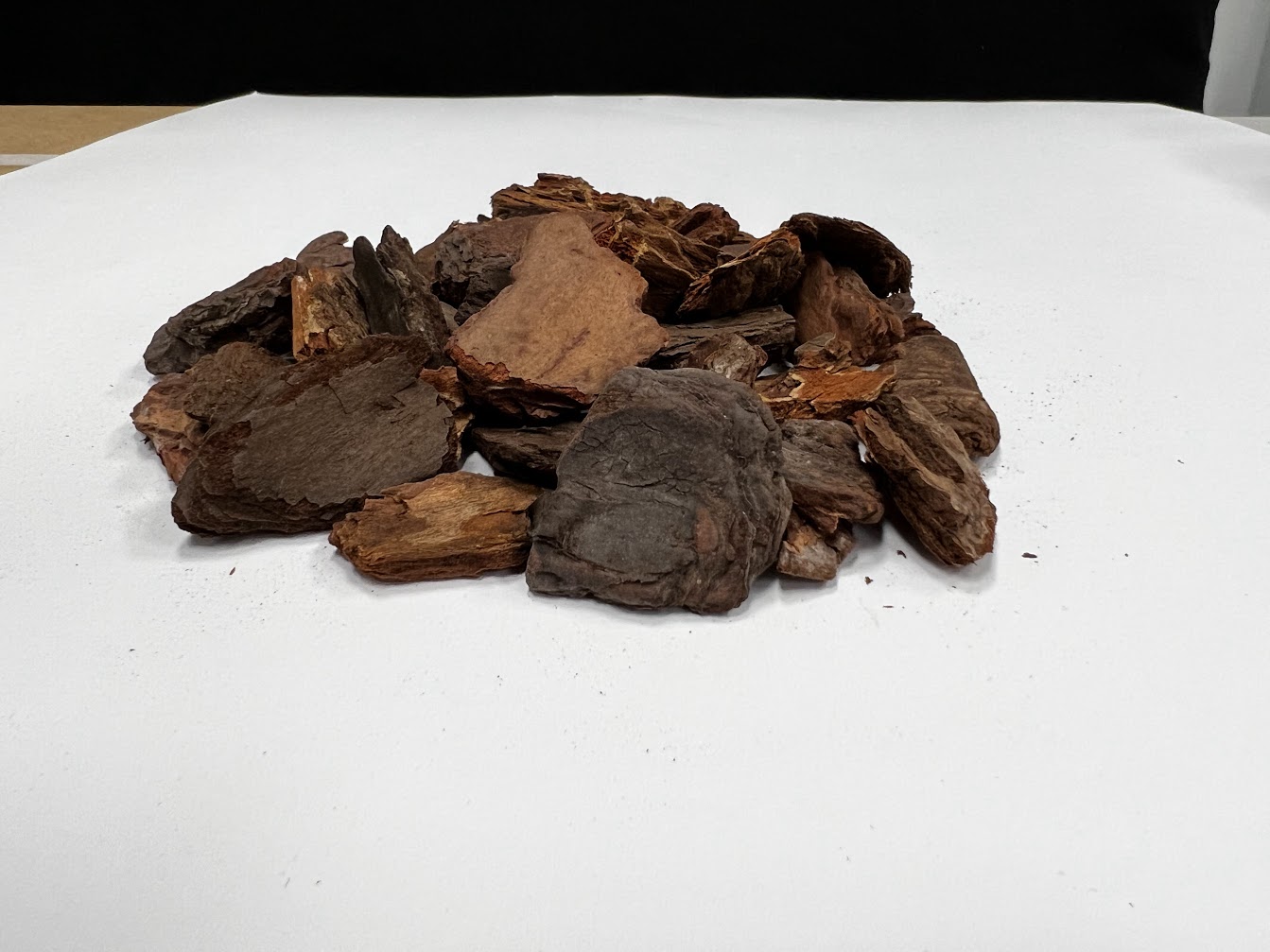}
         \caption{Wood Chips}
         \label{fig:hw_exp_wood_chips}
     \end{subfigure}
     \label{fig:hw_materials}
   \caption{Three types of real granular materials.}
\end{figure}

To show that our approach generalizes to different materials, we have chosen three different granular materials.
Our material selection consists of gravel (rounded edges), grit (sharp edges), and wooden chips.
The materials are shown in Figures \ref{fig:hw_exp_grind}, \ref{fig:hw_exp_grit}, \ref{fig:hw_exp_wood_chips}.

\subsection{Hardware Results}
Similar to the simulation experiments, we will demonstrate the functionality of our proposed planner in four different tasks that consist of gathering to a single target, separating into two clusters, separating into four clusters, and shaping the letters "E", "T", "H".
For the hardware experiments, we do not additionally randomize the target distribution as was done in the simulation experiments. Instead we use the respective \emph{nominal} target shapes shown in Figure \ref{fig:sim_exp_tgt_shapes}.

\begin{figure}[]
     \centering
     \begin{subfigure}[b]{0.49\columnwidth}
         \centering
         \includegraphics[width=\columnwidth, trim={0cm 0cm 0cm 0cm},clip]{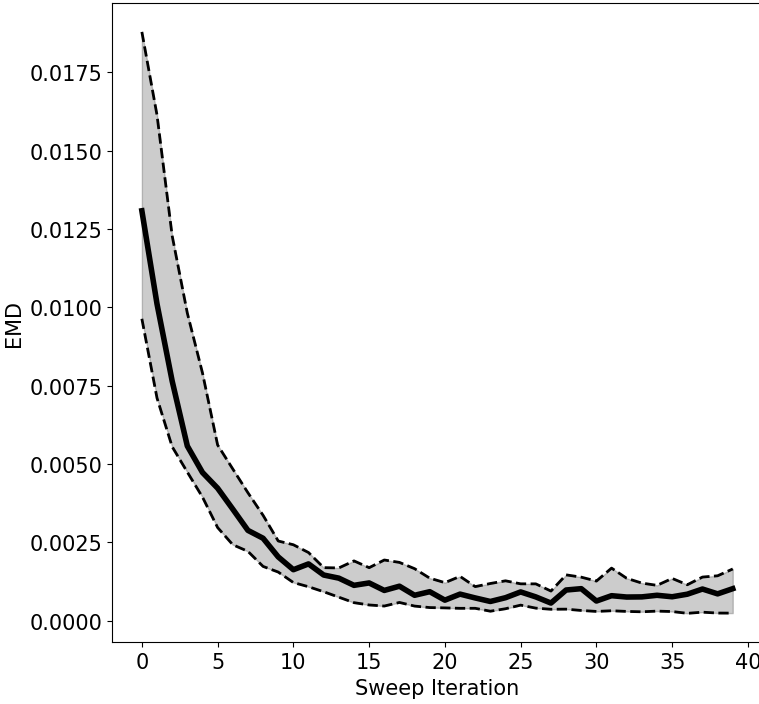}
         \caption{Quantiles for EMD.}
         \label{fig:hw_exp_summary_emd}
     \end{subfigure}
     \hfill
     \begin{subfigure}[b]{0.49\columnwidth}
         \centering
         \includegraphics[width=\columnwidth, trim={0cm 0cm 0cm 0cm},clip]{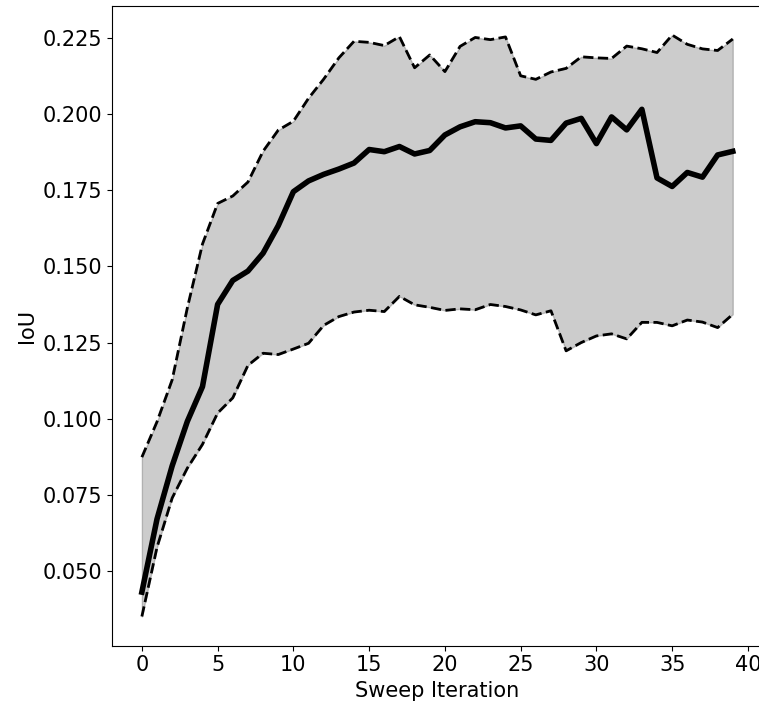}
         \caption{Quantiles for IoU.}
         \label{fig:hw_exp_summary_iou}
     \end{subfigure}
     \label{fig:hw_exp_summary}
      \caption{5\%/50\%/95\% quantiles for hardware experiments for gathering and separation.}
\end{figure}

However, we perform the three different tasks on \emph{all} three materials.
The results from these nine runs are summarized in the quantile plots showing the evolution of the median, and the $5\%$ as well as $95\%$ quantile curves for both the EMD as well as IoU in Figures \ref{fig:hw_exp_summary_emd}-\ref{fig:hw_exp_summary_iou}, which show consistent progress towards the target shape with small variance.
For the letter task, see Figure \ref{fig:hw_exp_eth_grit} for the qualitative results on shaping the letters "E", "T", and "H" with grit stones.

\section{CONCLUSIONS}
In this work, we proposed a material-agnostic planner for transforming piles of granular material from arbitrary initial shapes to arbitrary target shapes by means of sweeping motions with a robot arm. 
We leverage the mathematical framework of optimal transport and adapt it to sweep-based planning through a reactive next-best-sweep sampling-based approach and show that it is a promising avenue for granular material manipulation in robotics.
We have validated the approach in large-scale simulation and hardware experiments and compared them to two simpler baselines that tend to work on simpler tasks, like gathering, but fail at more difficult tasks that require more precise reasoning about how to efficiently re-distribute the material. We have shown that the proposed planner successfully generalizes to multiple input shapes, target shapes, and granular materials without requiring sophisticated or computationally expensive dynamics models of the materials.
Finally, while we assumed that the state of the granular material is given as a volumetric height map, this only works for larger particles that can be picked up by the depth sensor. For small-grained materials, like sand, one has to either use very high-precision depth sensors or leverage other sensor modalities to estimate the volume.

%%%%%%%%%%%%%%%%%%%%%%%%%%%%%%%%%%%%%%%%%%%%%%%%%%%%%%%%%%%%%%%%%%%%%%%%%%%%%%%%
% \section*{Acknowledgement}
\bibliographystyle{IEEEtran}
\bibliography{bibliography/references}

\end{document}